\documentclass[10pt,twocolumn,letterpaper]{article}

\usepackage[pagenumbers]{cvpr} 
\usepackage{colortbl}
\usepackage{graphicx}
\usepackage{multirow}
\usepackage{makecell}

\definecolor{cvprblue}{rgb}{0.21,0.49,0.74}
\usepackage[pagebackref,breaklinks,colorlinks,allcolors=cvprblue]{hyperref}

\def\paperID{1140} 
\def\confName{CVPR}
\def\confYear{2025}

\title{HumanDiT: Pose-Guided Diffusion Transformer \\ for Long-form Human Motion Video Generation
}

\author{
	Qijun Gan~\(^{1,2}\)\footnotemark[2]~
 \footnotemark[3]
	 \quad Yi Ren~\(^2\)\footnotemark[2]
	 \quad Chen Zhang~\(^2\)\footnotemark[2]
	 \quad Zhenhui Ye~\(^1\)
	 \quad Pan Xie~\(^2\)
	\\
        Xiang Yin~\(^2\)\footnotemark[4]
         \quad Zehuan Yuan~\(^2\)
         \quad Bingyue Peng~\(^2\)
         \quad Jianke Zhu~\(^1\)
	\\
	\(^1\)Zhejiang University \quad \(^2\)ByteDance  \\
	\texttt{\{ganqijun,jkzhu\}@zju.edu.cn},\\
	\texttt{\{ren.yi,yinxiang.stephen\}@bytedance.com}
    \\ \url{https://agnjason.github.io/HumanDiT-page/}
}

\begin{document}

\twocolumn[{%
    \maketitle
    \centering
    \vspace{-0.2in}\includegraphics[width=1\linewidth]{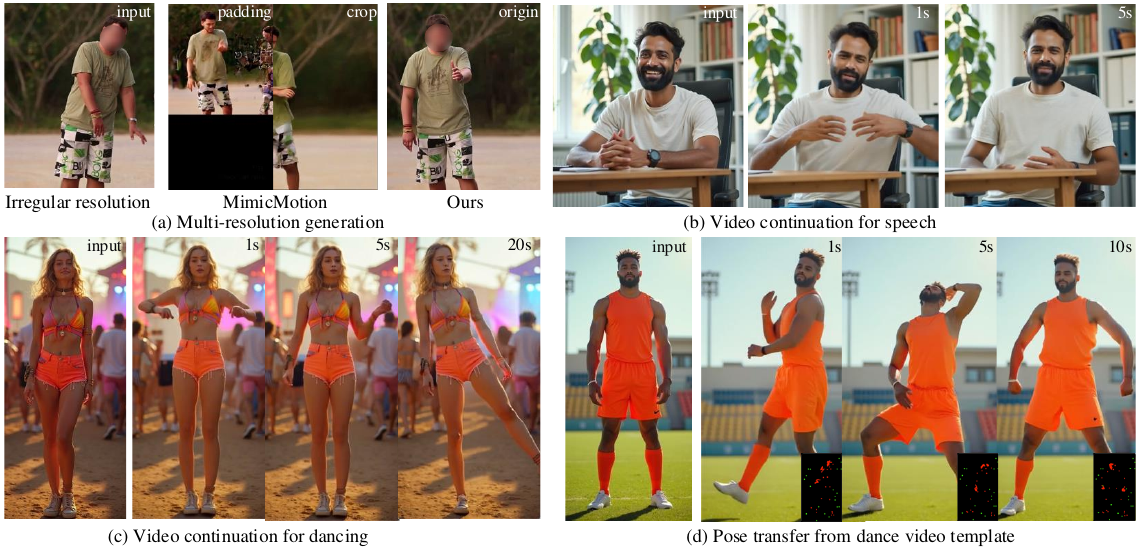}
    \vspace{-0.25in}
    \captionof{figure}{HumanDiT is a framework designed to generate high-fidelity, long human motion videos in diverse scenes and flexible resolution.}
    \label{fig:teaser}
    \vspace{0.10in} 
}]
\footnotetext[2]{These authors contributed equally to this work.} 
\footnotetext[3]{Done during an internship at ByteDance.} 
\footnotetext[4]{Corresponding author.} 
\footnote{Due to legal and portrait rights considerations, all real human faces in the paper are blurred, while unblurred images are generated by diffusion model.}

\begin{abstract}
Human motion video generation has advanced significantly, while existing methods still struggle with accurately rendering detailed body parts like hands and faces, especially in long sequences and intricate motions. Current approaches also rely on fixed resolution and struggle to maintain visual consistency. 
To address these limitations, we propose HumanDiT, a pose-guided Diffusion Transformer (DiT)-based framework trained on a large and wild dataset containing 14,000 hours of high-quality video to produce high-fidelity videos with fine-grained body rendering. Specifically, (i) HumanDiT, built on DiT, supports numerous video resolutions and variable sequence lengths, facilitating learning for long-sequence video generation; (ii) we introduce a prefix-latent reference strategy to maintain personalized characteristics across extended sequences.
Furthermore, during inference, HumanDiT leverages Keypoint-DiT to generate subsequent pose sequences, facilitating video continuation from static images or existing videos. It also utilizes a Pose Adapter to enable pose transfer with given sequences.
Extensive experiments demonstrate its superior performance in generating long-form, pose-accurate videos across diverse scenarios. 
\end{abstract}    
\section{Introduction}
\label{sec:intro}

The generation of realistic human motion videos has garnered significant attention in recent years, particularly with the advancements in generative artificial intelligence~\cite{hu2024animate,guo2023animatediff,chang2023magicdance,zhu2024champ,zhang2024mimicmotion,xu2024magicanimate,xue2024follow,lin2024cyberhost}. Compared to image generation, video generation presents greater challenges as it requires not only high-quality visuals but also smooth temporal consistency across frames to maintain visual coherence~\cite{zhang2024mimicmotion}. Applications such as virtual humans, animated films, and immersive experiences drive the need for reliable, high-quality motion video generation methods.


Despite notable advances~\cite{hu2024animate,zhu2024champ,xue2024follow,wang2024unianimate,xu2024magicanimate,chang2023magicdance,zhang2024mimicmotion,lin2024cyberhost}, current methods for generating human motion videos still face key limitations. 
First, achieving temporal consistency in long-sequence generation remains challenging due to multiple batches of inference.  Most models~\cite{hu2024animate,xue2024follow,lin2024cyberhost,zhang2024mimicmotion} restrict the number of frames per sequence with U-Net architectural~\cite{hu2024animate} constraints, and continuity methods~\cite{zhang2024mimicmotion,wang2024unianimate} based on overlap fail to prevent error propagation or ensure temporal consistency. While DiT-based method Human4DiT~\cite{shao2024human4dit} first introduces DiT in human video generation, they perform a single inference on a video with a fixed short duration and resolution.
Second, the ability to generalize across varied scenarios is limited, primarily due to the absence of extensive and diverse datasets. Additionally, achieving high-fidelity rendering of facial and hand details poses challenges, often resulting in blurred or inconsistent outputs, especially in extended sequences. 
Third, most of them rely on fixed-resolution inputs, requiring resizing or padding, which impacts both quality and flexibility, as illustrated in Figure~\ref{fig:teaser}(a). 
Finally, most approaches~\cite{wang2024disco,hu2024animate,zhang2024mimicmotion,xue2024follow} are specifically designed for pose transfer with a given poses sequence, and any misalignment in pose can result in visual artifacts.

To tackle these challenges, we present HumanDiT, an adaptable pose-guided body animation framework designed for diverse resolutions and long-form video generation (up to 20 seconds). We address the aforementioned challenges with an expert DiT for pose-guided video; a large-scale and high-quality dataset for diverse scenarios; an Keypoint-DiT for pose generation and a pose adapter for pose transfer.

First, to address variable resolution and dynamic sequence length, we replace the traditional U-Net diffusion model with the Diffusion Transformer (DiT)~\cite{peebles2023scalable}. Employing a prefix-latent reference strategy, HumanDiT preserves visual consistency across inputs while accommodating diverse resolutions and durations. The proposed pose guider facilitates the capture of temporal and spatial features via patch-based extraction, ensuring precise pose guidance. This DiT architecture enables sequence parallelism~\cite{li2021sequence}, optimizing training and inference for high-resolution, long-duration video generation.

Second, to enhance generalization and rendering quality, we gather a large-scale, diverse dataset comprising 14,000 hours of in-the-wild videos. This dataset is collected using a novel data processing pipeline, which includes a data extraction sequence and filtering strategies that employ scoring models to assess and select samples based on image clarity, particularly in detail-rich regions such as hands and teeth, thereby improving dataset reliability. We also incorporate text masks into the training process to prevent the model from learning text artifacts present in the wild data.

Third, we leverage an expert Keypoint-DiT for pose generation, thereby supporting a wide range of downstream applications beyond pose transfer~\cite{hu2024animate,zhang2024mimicmotion}. For video continuation, Keypoint-DiT enables the creation of longer, continuous, and natural pose sequences from a single initial pose, facilitating video continuation with free-flowing motion. 
For pose transfer, a pose adapter is introduced to align initial poses, which are subsequently refined by the Keypoint-DiT, thereby improving the accuracy of facial and hand details. In pose transfer, transitional frames are utilized for refinement to bridge the pose gap between the reference image and the guided poses.

In summary, our contributions are as follows: 1) We present HumanDiT, a DiT-based human body animation framework employing a prefix-latent strategy, which facilitates variable resolutions, dynamic sequence lengths, and high-fidelity video generation. 2) We develop a 14,000-hour in-the-wild video dataset with a structured data pipeline and a lightweight clarity scoring model to filter for high-quality videos. 3) We introduce a DiT-based pose generation module and pose adapter, enabling HumanDiT to support diverse applications such as long-form video continuation and pose transfer. 4) We achieve superior quantitative and qualitative performance across various scenes, outperforming state-of-the-art methods in comprehensive evaluations.

\section{Related Works}
\label{sec:relate}

\noindent\textbf{Image and Video Generation.} Recent years, image and video generation has made substantial advancements~\cite{vondrick2016generating, ge2022long, bar2023multidiffusion, zeng2024make}, beginning with early approaches based on GANs~\cite{goodfellow2014generative, chan2019everybody, ren2020deep, siarohin2019animating, siarohin2021motion} and VQVAE-based transformers~\cite{ge2022long, xu2016msr, sun2024autoregressive}, which laid the groundwork for synthesizing realistic videos. However, these methods often struggled with issues like temporal inconsistency and the high computational demands associated with modeling continuous motion. More recently, diffusion models (DMs)~\cite{song2020denoising, ho2020denoising, ni2023conditional} have emerged as a powerful alternative, offering greater stability and control. Unlike GANs, diffusion models progressively refine noisy inputs into coherent frames, leading to more robust and photo-realistic outputs. Latent Diffusion Models (LDM)~\cite{rombach2022high}, in particular, have optimized this process by performing it in lower-dimensional latent spaces. 

For video generation, these diffusion models~\cite{guo2023animatediff, bar2024lumiere} go beyond image synthesis by incorporating temporal layers and attention mechanisms to better model spatial-temporal relationships. To maintain temporal continuity, some approaches~\cite{singer2022make,blattmann2023stable, zhou2022magicvideo, wang2023dreamvideo} directly extend the 2D U-Net~\cite{ronneberger2015u}, pre-trained on text-to-image tasks into 3D. Techniques like Stable Video Diffusion (SVD)~\cite{blattmann2023stable} and AnimateDiff~\cite{guo2023animatediff} represent notable developments, as they extend 2D image generation frameworks to handle the additional complexity of continuous video segments. Recent methodologies~\cite{xu2024easyanimate, yang2024cogvideox, openai2024sora, zhang2024tora} have shown enhanced potential by integrating specialized motion modules and a DiT-based framework~\cite{peebles2023scalable}. Concurrently, 3D Variational Autoencoders (VAEs)~\cite{kingma2013auto} alleviate the computational demands of video data processing through advanced compression techniques. 

\noindent\textbf{Human Animation.} Pose-guided human image animation has seen significant advancements, particularly with the incorporation of pose estimation methods such as OpenPose~\cite{cao2017realtime}, DWpose~\cite{yang2023effective} and Sapiens~\cite{khirodkar2024sapiens} for guiding motion synthesis. Early approaches~\cite{liao2020speech2video, weng2022humannerf, ginosar2019learning, ye2023geneface}, often based on GANs~\cite{goodfellow2014generative} or NeRFs~\cite{mildenhall2021nerf}, focused on transferring poses between images using explicit skeleton representations, while these struggled with temporal consistency and flexibility in motion transfer. DisCo~\cite{wang2024disco} integrates character features using CLIP and incorporates background features through ControlNet~\cite{zhang2023adding}.

The rise of diffusion models (DMs) has greatly improved the quality of both image and video generation. Recent methods, like MagicAnimate~\cite{xu2024magicanimate} and Animate Anyone~\cite{hu2024animate}, introduce specialized motion modules and lightweight pose guiders to ensure accurate pose-to-motion transfer. Champ~\cite{zhu2024champ} relies on parametric models SMPL~\cite{loper2023smpl}, which offer realistic human representations for pose and shape analysis. Based on 4D diffusion transformer, Human4DiT~\cite{shao2024human4dit} generates free-view dynamic human videos with SMPL sequences and camera parameters. UniAnimate~\cite{wang2024unianimate} utilizes first frame conditioned input for consistent long video generation. Mimicmotion~\cite{zhang2024mimicmotion} employs cross-frame overlapped diffusion to generate extended animated videos. Xue~\textit{et al.}~\cite{xue2024follow} utilize an optical flow guide to stabilize backgrounds and a depth guide to handle occlusions between body parts. Cyberhost~\cite{lin2024cyberhost} enhances hand and face generation with a Region Codebook Attention mechanism. Moreover,  Tango~\cite{liu2024tango} synthesizes co-speech body-gesture videos by retrieving matching reference video clips. Animate-X~\cite{tan2024animate} extends beyond human to anthropomorphic
characters with various body structures by utilizing implicit and explicit pose indicator.

For precise rendering of hands and faces, HandRefiner~\cite{lu2024handrefiner} employs the ControlNet module to correct distorted hands. ShowMaker~\cite{yangshowmaker} utilizes latent features aligned with facial structures for enhancement. RealisDance~\cite{zhou2024realisdance} and TALK-Act~\cite{guan2024talk} leverage 3D priors of hands and faces, providing accurate 3D or depth information as conditioning inputs.
\begin{figure*}[t]
    \centering
    \includegraphics[width=1.0\linewidth]{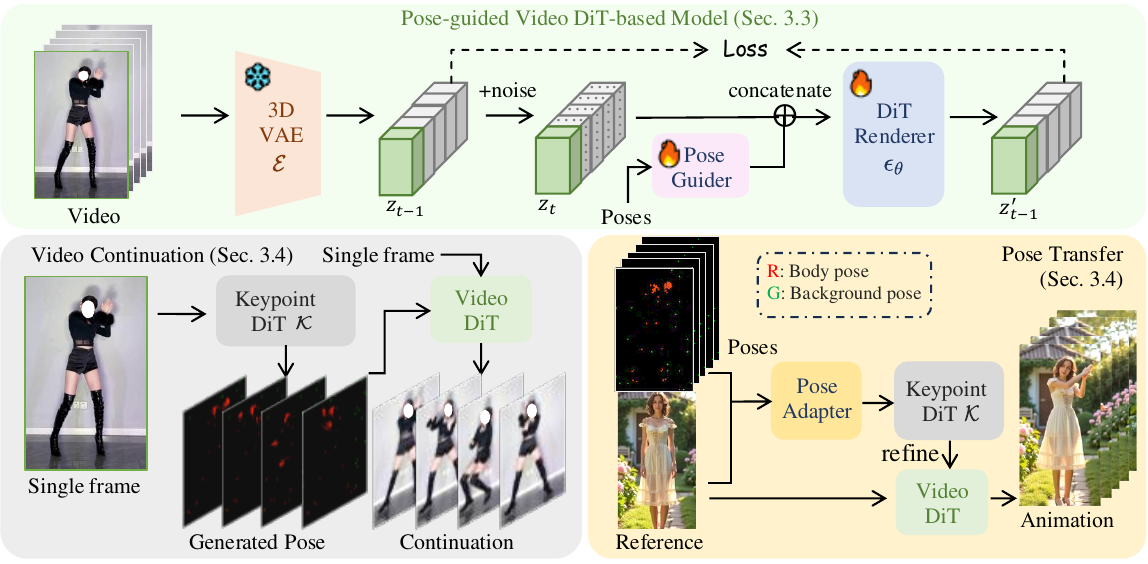}
    \vspace{-0.25in}
    \caption{\textbf{The overview of HumanDiT.} HumanDiT focuses on generate videos from a single image using a pose-guided DiT model. A 3D VAE is employed to encode video segments into latent space. With 3D full attention, the initial frame (green border) serves as a noise-free prefix latent (green cube) for reference. The pose guider extracts body and background pose features, while the DiT-based denoising model renders the final pixel results. During inference, the keypoint-DiT model produces subsequent motions based on the pose of the first frame. With a guided pose sequence, the pose adapter transfers and refines poses via keypoint-DiT to animate the reference image.}
    \label{fig:framework}
    \vspace{-0.15in}
\end{figure*}
\section{Methodology}
\label{sec:method}


\subsection{Preliminaries}
\label{subsec:pre}

\noindent\textbf{Latent Diffusion Models.} The latent diffusion models~\cite{rombach2022high} learns a denoising process to simulate the probability distribution within the latent space. To reduce the computational load, the image $x$ is transformed from pixel space into a latent space feature $z_0=\mathcal{E}(x)$ with a Variational Autoencoder (VAE) Encoder~\cite{kingma2013auto} $\mathcal{E}$. During the forward diffusion process, Gaussian noise is iteratively added to $z_0$ at various timesteps $t$ to $z_t$ as
$
    q(z_t | z_{t-1}) = \mathbb{N}(z_t;\sqrt{1-\beta_t}z_{t-1}, \beta_tI),
$
where $\beta$ represents a sequence schedule. The denoising process is defined as a iterative Markov Chain that progressively denoises the initial Gaussian noise $z_T \in \mathcal{N}(0, I)$ into the clean latent space $z_0$. The denoising function of LDM is commonly implemented with U-Net~\cite{ronneberger2015u} or Transformers~\cite{vaswani2017attention,peebles2023scalable}, which is trained by minimizing the mean square error loss as
$
    \mathcal{L} = \mathbb{E}_{z_t,c,t,\epsilon \sim \mathcal{N}(0,I)}[||\epsilon - \epsilon_\theta(x_t;c,t)||^2_2],
$ where $\epsilon_\theta$ represents parameterized network for predicting noise and $c$ denotes an optional conditional input. Subsequently, the denoised latent space feature is decoded into image pixels using the VAE Decoder $D$.

\noindent\textbf{Diffusion Transformer (DiT).} 3D U-Net is utilized widely in human video generation~\cite{hu2024animate,zhu2024champ,lin2024cyberhost,xue2024follow} with temporal attention modules to generate continuous video segments, while they always rely on fixed input resolution and reference network to control personality consistency, which cost additional calculations. The DiT~\cite{peebles2023scalable} combines the strengths of diffusion model with transformers~\cite{vaswani2017attention}, which address the limitations of U-Net-based LDM. By leveraging patchify~\cite{peebles2023scalable,yang2024cogvideox} and Rotary Position Encoding (RoPE), denoising model $\epsilon_\theta$ is able to process diverse image resolution and sequence length. 
RoPE is a positional encoding method that incorporates relative positional information into the embedding space by applying rotational transformations. Unlike traditional fixed or absolute encodings, RoPE allows the model to capture relative positional relationships in a resolution-agnostic manner, enabling it to generalize effectively across inputs of varying dimensions.

\subsection{Data Preparation}
\label{subsec:data}

We develop a structured data processing pipeline, which collects millions of videos that capture a wide range of human motions, including speeches, dancing, movies, and daily scenarios. To better understand human motions, a large amount of data is essential.

For each video sequence, the data samples consist of four components: (1) image sequences containing various human motions, (2) corresponding body pose sequences, (3) background keypoint sequences and (4) text areas. First, Yolo~\cite{Jocher_Ultralytics_YOLO_2023} is employed to track and distinguish different individuals in the video and crop them accordingly. To prevent overlapping bodies interfering with the keypoint estimation, cropped videos featuring multiple people are excluded. Next, the body poses sequences are labeled with Sapiens~\cite{khirodkar2024sapiens}, a strong tool for robust and stable pose estimation. The background keypoints are extracted by CoTracker~\cite{karaev2023cotracker}. We distinguish the foreground and background based on the human mask from the first frame. Additionally, since subtitles in the video can lead to undesired text generation during inference, PaddleOCR~\cite{PaddleOCR} is utilized to recognize the text bounding boxes. Finally, to ensure our data quality, those videos are filtered out where wrists or most of the body are not visible. For video clarity assessment, we manually label the clarity of 150K sets of teeth and hand images, each set containing five images with corresponding clarity ranks. Then a lightweight model~\cite{he2016deep} is employed to evaluate the clarity of hands and teeth, filtering high-quality videos based on the clarity score. More details are provided in Appendix.A.

With the data processing pipeline, a large-scale human video dataset is obtained. All videos are split into segments no longer than 20 seconds, and the final dataset consists of 4.5 million video clips, totaling 14,000 hours of duration.

\subsection{Pose-guided Video DiT Model} 
\label{subsec:dit}

Given a reference image $x^{0}$, the primary objective of HumanDiT is to generate high-quality, continuous human motion videos with consistent visual appearance. This task involves synthesizing realistic human motion and camera movements that mimic the pose sequence, while maintaining visual fidelity to the input image. Recent methods have shown promising generation capacity, while they always fix the image resolution and sequence length and require two stage training. To overcome this constraint, as shown in Figure~\ref{fig:framework}, HumanDiT employs scalable diffusion transformer (DiT)~\cite{peebles2023scalable} as its foundational model. With Rotary Position Embedding (RoPE)~\cite{su2024roformer} and patchify, DiT $\epsilon_\theta$ can handle videos of diverse size and length.

Following CogVideoX~\cite{yang2024cogvideox}, HumanDiT adopts a pre-trained 3D VAE with a video compression module. It incorporates 3D convolutions to compress videos both spatially and temporally, which generates longer videos while maintaining the quality and continuity. Specifically, given a video $\mathbf{x}:\{x^0,x^1,...,x^{4f+1}\}$ of size $(4f + 1) \times 8h \times 8w \times 3$, the video latent vector $\mathbf{z}_0$ can be encoded by the 3D VAE $\mathbf{z}_0 = \mathcal{E}(\mathbf{x})$, where $\mathbf{z}_0 \in \mathbb{R}^{(f+1) \times h \times w \times s}$ and $s$ denotes length of latent channel.

\noindent\textbf{Reference Strategy.} Visual consistency has always been a challenge in video generation. Unlike previous works~\cite{hu2024animate,zhang2024mimicmotion}, HumanDiT does not rely on a reference network with the same structure as the denoising model to transfer reference features. Instead, we simply employ a prefix-latent reference strategy. Thanks to the 3D full attention design in DiT, the model can directly capture reference image features from the prefix latent. Specifically, the latent feature vector of a video segment is denoted as $ \mathbf{z}_0 = \{z^0, z^1, z^2, \dots, z^f\} $. During the forward diffusion process, noise is gradually added to $ \{z^1, \dots, z^f\} $, while the prefix latent $ z^0 $ serves as a reference for the model in capturing the input image characteristics. During training, the predicted noise $ z_{t-1}^0 $ of $ z_t^0 $ is excluded from loss computation. Since the first feature $z^0$ of the 3D VAE is obtained from the first frame $x^0$ of the image independently, during inference, the prefix latent is obtained via the encoder $ z^0 = \mathcal{E}(x^0) $.

\noindent\textbf{Pose Guidance.} To extract pose-guided features, inspired by patchify, we introduce a linear pose guider for HumanDiT. Previous approaches often utilize multi-layer convolutional blocks, which effectively capture the contextual information of pose images, while these methods fail to account for temporal features. Unlike these methods, our proposed pose guider converts spatial pose pixels into a sequence of tokens by applying patchify. 

Specifically, the pose images are composed of pixel points representing the positions of body keypoints and background keypoints, where these keypoints are set to 1 and the rest to 0. The pose images corresponding to the video are denoted as  $
    \mathbf{P} \in \mathbb{R}^{(4f+1)\times4h\times4w\times d},
$ where $d$ represents the pose image dimension. To ensure patchify effectively and prevent future information from influencing present or past predictions~\cite{yang2024cogvideox}, padding is applied at the beginning of the pose images for $3$ frames. Subsequently, $ \mathbf{P} $ is patchified across the temporal, height, and width dimensions according to the patch size. To ensure alignment with the latent features $ \mathbf{z} $, the patch size is fixed at 4. Following the patchification process, the resulting pose tokens are represented as $  \mathbf{P}' \in \mathbb{R}^{(f+1) \times h \times w \times 64d} $, which can be projected into the condition $c$ in latent space, effectively capturing both spatial and temporal characteristics. This transformation is implemented through a linear layer and trained jointly with DiT model.

Skeletal representations tend to vary in size across resolutions, which can lead to inconsistencies and overlapping points in pose conditioning. To address this, HumanDiT uses pixel-based representations, which effectively capture pose information while reducing overlap. Heat-map-based conditioning~\cite{wang2021one} typically requires substantial memory due to the large number of body keypoints and extended sequences. To optimize memory usage and minimize pixel overlap, we set the pose image dimension  $d = 8$ , where the first 7 dimensions encode body keypoints and the last dimension is drawn with no more than 20 background keypoints. Each body keypoint is sequentially mapped to one of the first 7 dimensions in a round-robin fashion.

\noindent\textbf{Text Mask.} Subtitles in videos negatively impact the training and inference of HumanDiT. 
To address this issue, we implement a text-masking strategy to control the feature learning scope during training. Specifically, during the training process, text masks are applied to the regions corresponding to text in the latent feature space, preventing the model from calculating loss for those areas. This approach ensures that the model focuses on learning relevant visual and temporal features. Additionally, to enhance the model’s understanding that text regions should not be learned, the text mask is also provided as a conditional input to the model. This allows the model to explicitly recognize and ignore the masked regions during training and concentrate on human motions without being influenced by subtitles.


\subsection{Progressive Long Video Generation for Human}
\label{subsec:infer}

The generation of long continuous video sequences has been a challenge due to computational resource constraints. To address this, HumanDiT simply takes the final frame of one segment as the initial frame of the next with the prefix-latent strategy. Generated frames are consistent with the input image, which ensures visual continuity in the generated video. Additionally, with the transformer architecture and our designed pose guider, sequence parallelism~\cite{li2021sequence} is applied along the temporal dimension, allowing computations to be distributed across multiple devices. Longer video sequences is able to be trained and inferred efficiently. 


\noindent\textbf{Video Continuation from single image.}
To enable portrait animation from a single image, we also design a keypoint generation module that adopts the similar architecture as DiT, which generating keypoint sequences based on the pose in a given image. Specifically, given the initial keypoint $ j_0 $ from an image, the Keypoint-DiT $ \mathcal{K} $ is used to iteratively denoise and predict the keypoints for the subsequent $ m $ frames, expressed as 
$
    \{ j_1, j_2, \dots, j_m \} = \mathcal{K}(j_0).
$ Similar to Video DiT, Keypoint-DiT uses the pose from the first frame as a prefix pose and denoises the subsequent $t-1$ noise frames to generate the following motions.

With this pose generation module, HumanDiT can extend a single human pose into a dynamic and realistic motion. By predicting keypoint sequences, it enables our model to synthesize continuous motion without requiring full pose sequences as reference, making it adaptable to various video generation tasks.

\noindent\textbf{Pose Transfer with pose sequence.}
The U-Net-based LDM utilizes a Reference Network to incorporate reference images into the denoising process through cross attention. These reference images represent diverse scene and pose, making pose alignment with the reference images less critical. However, incorrect body proportions between the reference image and the input pose may lead to inaccurate visual results. In contrast, since HumanDiT generates human motion videos based on continuous poses, pose alignment is crucial. The main idea of our pose adapter is to decouple the motion sequence from the template and apply it to the skeleton of the initial reference image $ x^0 $.

Formally, the human pose action template is represented as a sequence $\mathcal{J} = \{j_k\}^m_{k=1} $, where $\mathcal{J}$ contains the position information of $m$ frames. The human pose in the reference image is denoted as $ j_0 $, which is consist of joints. Given a series of human joint pairs, the Euler distance of each joint $l_k$ and $l_0$ relative to its preceding joint can be calculated as $l^i = \| j^i - j^{i-1} \|_2$, where $j^{i-1}$ and $j^i$ are the neighboring keypoints. Since the human body moves in three-dimensional space, the distances in image space do not accurately reflect the actual joint lengths, which could mislead the transfer of the keypoints. To address this, the pose adapter traverses action sequences to obtain the maximum distance $l^{\prime}$ for each joint pair, using it as the reference joint length in the template pose. Given an initial joint, the remaining joint positions in $j_k$ can be transferred as the following
$
\hat{j}^i_k = \hat{j}^{i-1}_0 + l^i_0 \cdot (j^i_k - j^{i-1}_k)/l_i^{\prime}.
$

This procedure models the transformation of joint orientations and length variations within the template pose sequence, initiating calculations from the neck joint. Subsequent joint positions are derived step by step, ensuring that the directions and lengths of body parts remain consistent with those in the template video and resulting the aligned pose sequence $\hat{\mathcal{J}} = \{\hat{j}_k\}^m_{k=1}$ . 

\begin{table*}[htp!]
\centering\small

\begin{tabular}{c|ccccc|cc|cc}
\Xhline{1pt}
\textbf{Model} & \textbf{FID}  $\downarrow$   & \textbf{SSIM}   $\uparrow$ & \textbf{PSNR}   $\uparrow$  & \textbf{LPIPS}  $\downarrow$  & \textbf{L1}    $\downarrow$  & \textbf{FID-VID}  $\downarrow$  & \textbf{FVD}  $\downarrow$     & \textbf{Resolution} & \textbf{Frame} \\ \hline
DisCo~\cite{wang2024disco}       & 210 & 0.636              &  12.9& 0.439             & 6.29E-05     & 203      & 1808              & $256\times256$ & 1  \\ 
MagicDance~\cite{chang2023magicdance}          & 103 & 0.728              & 17.4& 0.327              &  3.43E-05  & 93.4         & 918   & $512\times512$ & 1 \\ 
AnimateAnyone~\cite{hu2024animate}         & 79.6&  0.737    &  18.1& 0.279              & 2.80E-05        & 68.8   & 730   & $576\times1024$ & 24 \\ 
MimicMotion~\cite{zhang2024mimicmotion}            & 65.4&  0.776              & 20.1& 0.238              &  2.76E-05  & 39.5         & 458   & $576\times1024$ & 72 \\ 
\cellcolor[HTML]{D9D9D9}{Ours}           & \cellcolor[HTML]{D9D9D9}{49.4} & \cellcolor[HTML]{D9D9D9}{0.838}             & \cellcolor[HTML]{D9D9D9}{22.8}  & \cellcolor[HTML]{D9D9D9}{0.186}           & \cellcolor[HTML]{D9D9D9}{1.73E-05}  & \cellcolor[HTML]{D9D9D9}{25.5}         & \cellcolor[HTML]{D9D9D9}{320}   & \cellcolor[HTML]{D9D9D9}$w,h<=1280$ & \cellcolor[HTML]{D9D9D9}25-249 \\ \Xhline{1pt}
\end{tabular}
\vspace{-0.1in}
\caption{Quantitative comparison on the TikTok dataset~\cite{jafarian2021learning} and our self-collected talking and dancing test datasets with existing pose-guided human body animation methods. Further details are provided in the supplemental material (Appendix.D.1).}
\label{tab:metrics}
\vspace{-0.1in}
\end{table*}

\begin{figure*}
    \centering
    \includegraphics[width=1.0\linewidth]{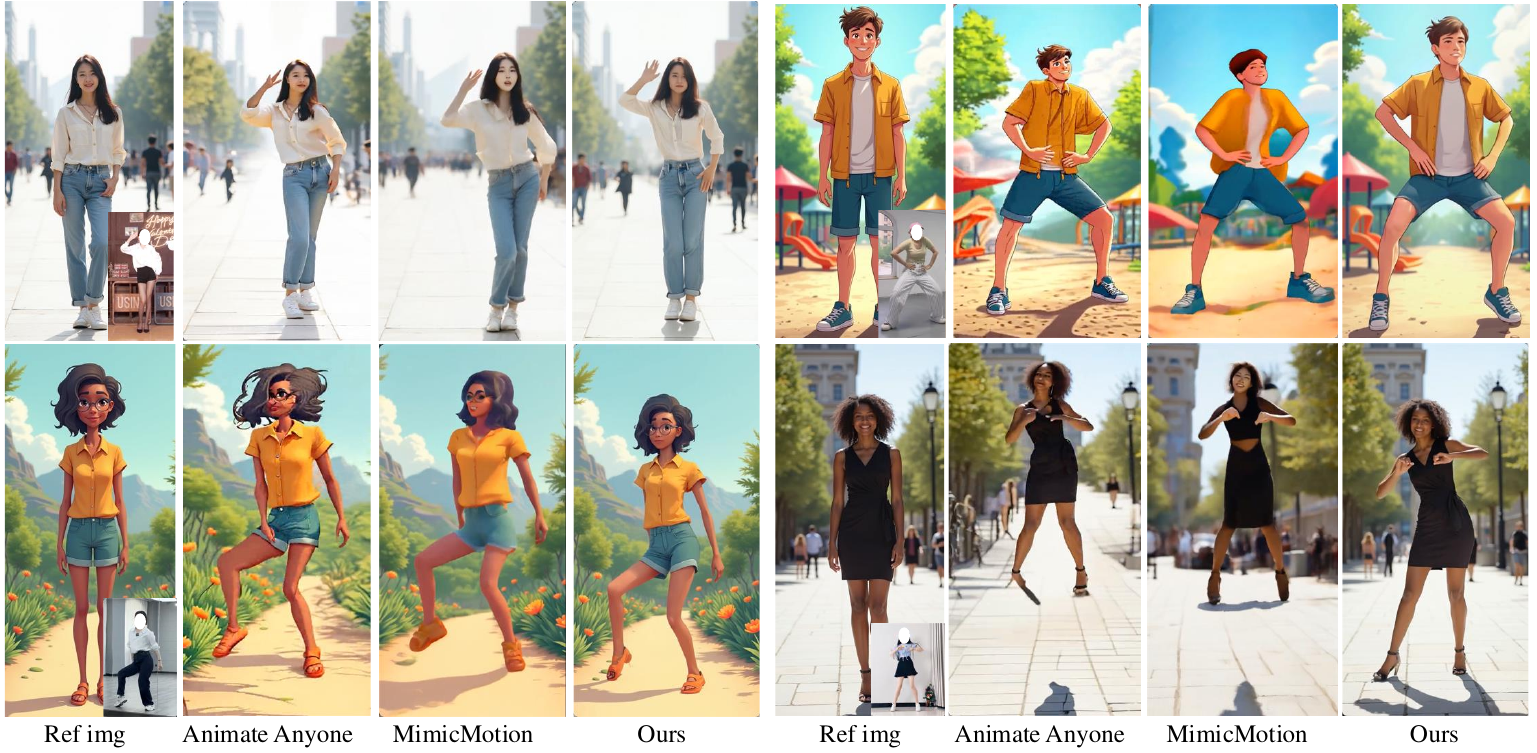}
    \vspace{-0.25in}
    \caption{Qualitative comparison. Our approach outperforms others in rendering quality and pose accuracy. }
    \vspace{-0.2in}
    \label{fig:tiktok}
\end{figure*}

\noindent\textbf{Pose Refinement.} Our DiT renderer $ \epsilon_\theta $ performs denoising based on continuous motion sequences. However, directly using $\{j_0, \hat{j_1}, \dots, \hat{j_m}\}$ as conditions for generation leads to a lack of smooth transition between $ j_0 $ and $ \hat{j_1} $. This gap results in visual artifacts such as motion blur and ghosting, particularly affecting the quality of the initial segment of the sequence. Moreover, the hand and face regions pose additional challenges compared to the body skeleton. Due to the complex articulation, as well as the absence of 3D motion information in 2D keypoints, pose transfer struggles to adequately capture facial and hand movements. 

To produce smooth transitional poses and enhance the alignment of facial and hand features with the reference image, we introduce a refinement module based on Keypoint-DiT $ \mathcal{K} $. Specifically, the pose-aligned sequence $ \hat{\mathcal{J}} $ is employed as a condition for Keypoint-DiT $ \mathcal{K} $. To ensure smooth transitions, $ \tau $ frames are padded between $ j_0 $ and $ \hat{j_1} $, forming the final motion sequence $ \mathcal{J}_{\text{fix}} = \mathcal{K}(j_0, \hat{\mathcal{J}},\tau) $. This sequence consists of the initial frame $ j_0 $, $ \tau $ transitional frames, and the aligned pose sequence $\hat{\mathcal{J}}$ of $ m $ frames. The refinement module effectively bridges the gap between $ j_0 $ and $ \hat{j_1} $, maintaining temporal continuity. Additionally, the pose and shape of the face and hands in $ \mathcal{J}_{\text{fix}} $ are refined based on observations from $ j_0 $, resulting in more accurate and realistic movements. 


\section{Experiments}
\label{sec:exp}

\begin{figure*}[htp]
    \centering
    \includegraphics[width=1.0\linewidth]{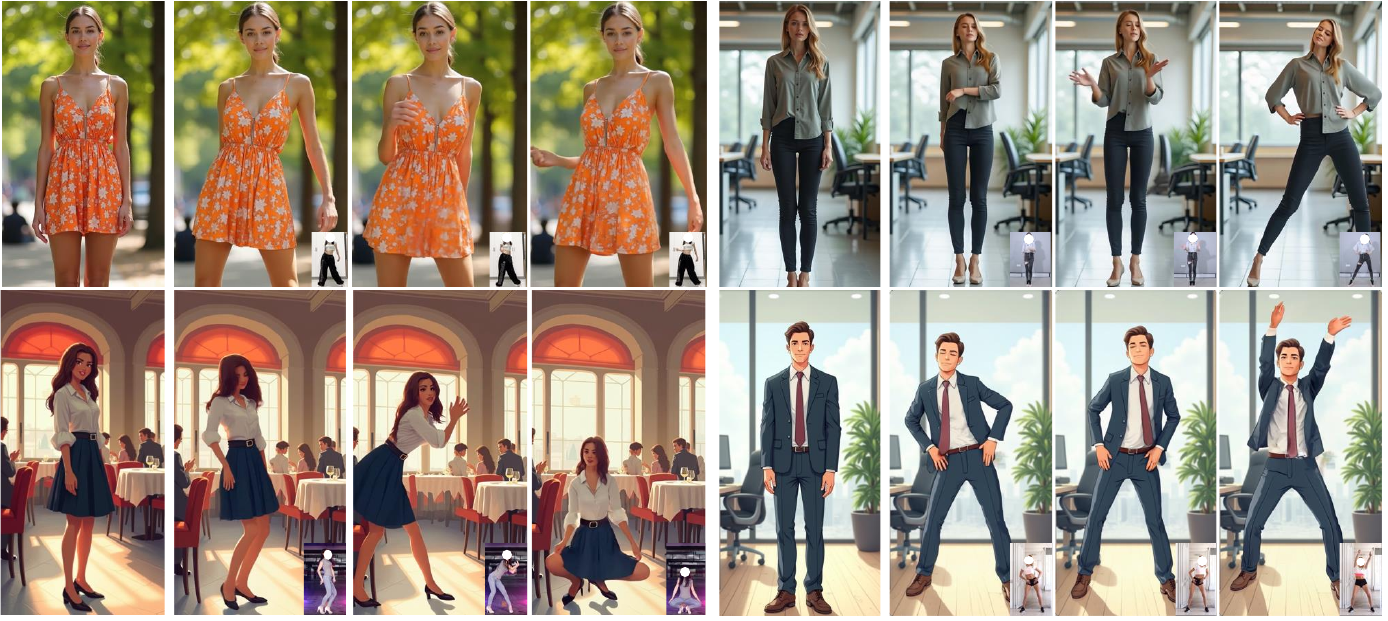}
    \vspace{-0.25in}
    \caption{The template pose-driven human rendering results of HumanDiT on the Flux~\cite{flux} generated images.}
    \vspace{-0.2in}
    \label{fig:templ}
\end{figure*}

\begin{figure}[htp]
    \centering
    \includegraphics[width=1.0\linewidth]{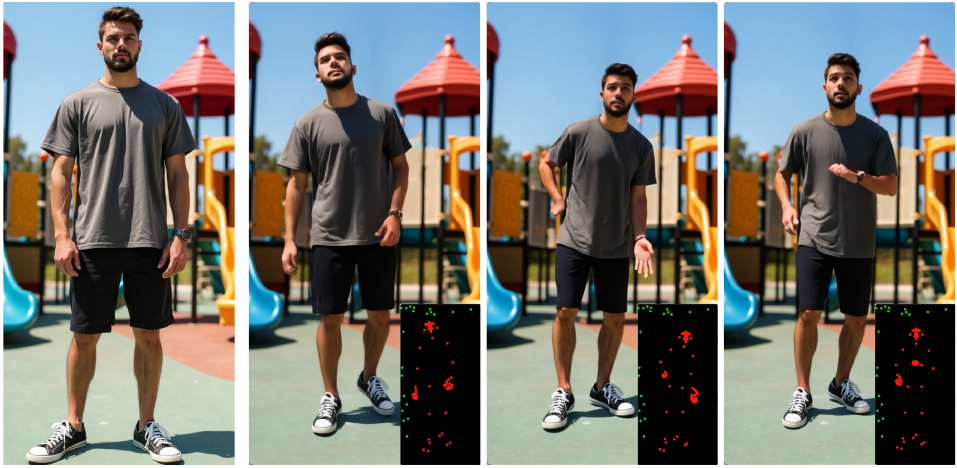}
    \vspace{-0.25in}
    \caption{The video continuation with generated motions.}
    \vspace{-0.15in}
    \label{fig:gene}
\end{figure} 

\subsection{Implementation Details}

Without the constraints of the reference network, the video DiT requires only a single-stage end-to-end training process. We adopt 3D VAE and DiT-based denoising model from CogVideoX-5b~\cite{yang2024cogvideox}. The weights of 3D VAE keep frozen throughout training. The model is trained on our own collected dataset, with max video resolutions capped at $1280$ and a frame rate of $12.5$ fps. To support dynamic resolution, we incorporate Rotary Position Embedding (RoPE)~\cite{su2024roformer} in the model, which allows for flexible adaptation with a relative positional
encoding to variable input sizes without retraining. During training, we expose the model to a range of resolutions, enabling it to learn effectively across diverse video qualities and sizes. To enable training on longer video sequences, we implement sequence parallelism, allowing each sequence to be trained across up to 8 GPUs. The maximum latent feature token size is constrained to ensure $(f+1)\times h\times w<480,000$. For a 720p video at $720\times1280$ resolution with a latent size of $90\times160$, the maximum latent sequence length is $33$, equivalent to $129$ original frames. The model is trained on 128 NVIDIA H100 GPUs (80GB) with a learning rate of $5e-5$. During inference, the classifier-free guidance (CFG) scale for the pose control is set to $1.5$.

\subsection{Results}

For quantitative evaluation, we select 30 video clips outside the training set and adopt sequences 335-340 from the TikTok dataset~\cite{jafarian2021learning}, following~\cite{zhang2024mimicmotion}, as the test set.

We compare our approach against state-of-the-art pose-guided human video generation methods, including Disco~\cite{wang2024disco}, MagicDance~\cite{chang2023magicdance}, AnimateAnyone~\cite{hu2024animate}, and MimicMotion~\cite{zhang2024mimicmotion}. For AnimateAnyone, we utilize the open-source implementation provided by MooreThreads~\cite{moore}. Consistent with \cite{wang2024disco}, we evaluate image quality using L1 error, Peak Signal-to-Noise Ratio (PSNR)\cite{hore2010image}, Structural Similarity Index Measure (SSIM)\cite{wang2004image}, Learned Perceptual Image Patch Similarity (LPIPS)\cite{zhang2018unreasonable}, and Fréchet Inception Distance (FID)\cite{heusel2017gans}. Video-level FID (FID-VID)\cite{balaji2019conditional} and Fréchet Video Distance (FVD)~\cite{unterthiner2018towards} are employed to assess video fidelity.

\noindent\textbf{Quantitative Evaluation.} Table~\ref{tab:metrics} presents a quantitative comparison of various methods evaluated on the TikTok dataset~\cite{jafarian2021learning} and our own collected dancing and talking scene videos. The results demonstrate that our proposed method outperforms all existing approaches, achieving notably lower L1 loss, higher PSNR and SSIM scores, and reduced LPIPS, FID-VID, and FVD values. This performance is attributed to the scale and diversity of our dataset, which enables HumanDiT to generate animated human videos across a wide range of scenes and conditions. 

Leveraging the architecture of DiT and RoPE, our model accommodates variable input resolutions and arbitrary sequence lengths, thereby providing flexibility in handling input images. In contrast, existing methods require center cropping or resizing with padding to fixed dimensions, which can compromise the ability to fully capture the entire image context. Additionally, in terms of video quality, MimicMotion~\cite{zhang2024mimicmotion} relies on overlapping frames between denoised segments to ensure smoother transitions. While our method achieves superior continuity by utilizing only the last frame of the preceding segment, simplifying the process while improving video performance.

\noindent\textbf{Qualitative Analysis.} We conduct qualitative comparisons between the selected baselines and our method across various scenarios. Figure~\ref{fig:tiktok} presents visual results of our test set, where the first frame of the video is used as the reference image. Compared to other methods, our approach demonstrates superior performance in pose simulation and visual consistency. Even when body parts in the reference image, such as hands, are blurry or obscured, our method effectively renders clear and detailed body parts. While MimicMotion also demonstrates strong pose-following capabilities, it struggles with preserving clothing details, often rendering overly smooth and inconsistent appearances. In contrast, our method generates vivid videos with pose guidance while maintaining visual consistency.


\noindent\textbf{Human Video Generation.} Figures~\ref{fig:templ} and ~\ref{fig:gene} demonstrate that our method effectively animates the reference human body using either a template or a generated pose sequence, as illustrated in the bottom right corner. Although our DiT-based model requires aligned and continuous poses, the proposed pose alignment strategy allows HumanDiT to animate reference images smoothly in synchronization with motion videos. Our approach produces high-quality, realistic character details while maintaining visual consistency with the reference image and ensuring temporal continuity across frames. The reference images depicted in the figures are generated by Flux~\cite{flux}. 

\subsection{Ablation Study}

\begin{figure}[htp]
    \centering
    \includegraphics[width=1.0\linewidth]{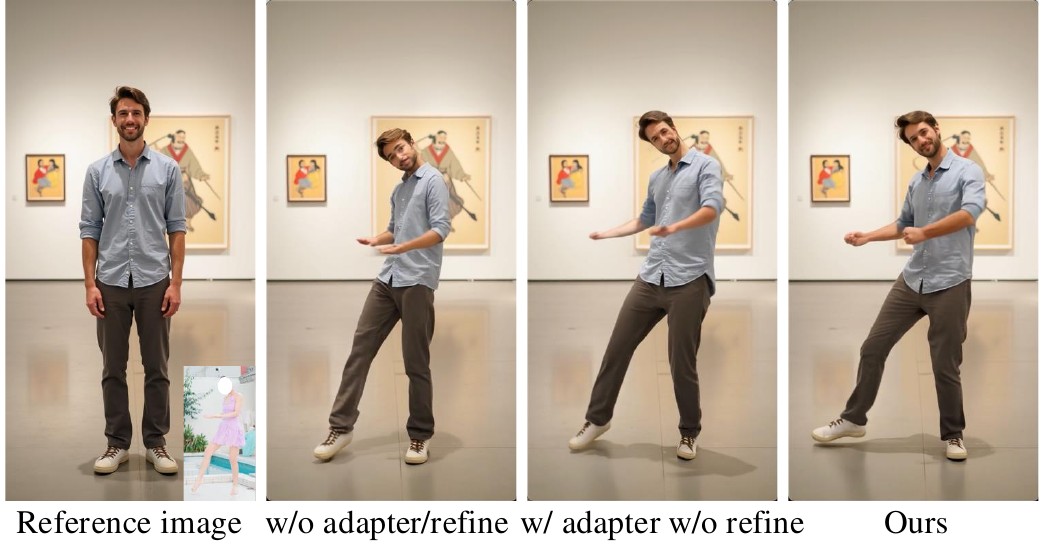}
    \vspace{-0.25in}
    \caption{Ablation study of pose adapter and pose refinement.}
    \vspace{-0.15in}
    \label{fig:abl}
\end{figure}

\begin{table}[t]
    \centering
    \footnotesize
    \begin{tabular}{p{0.3cm}cc|cccc}
    \Xhline{1pt}
        \textit{Size} & \textit{Tokens} & \textit{Ref.} & SSIM$\uparrow$ & PSNR$\uparrow$  & FID-VID$\downarrow$  & FVD$\downarrow$  \\ \hline
        5B & 80K & concat & 0.702 & 18.71 & 35.5  & 395 \\
        5B & 80K & prefix & 0.719 & 19.50 & 33.5 & 382 \\ \hline
        5B & 400K & prefix & 0.837 & \cellcolor[HTML]{D9D9D9}22.79 & 26.9 & 353 \\
      5B & 480K &
        prefix & \cellcolor[HTML]{D9D9D9}0.838 & 22.75 & \cellcolor[HTML]{D9D9D9}25.5 & \cellcolor[HTML]{D9D9D9}320 \\ \Xhline{1pt}
    \end{tabular}
    \vspace{-0.1in}
    \caption{Quantitative results of ablation study on maximin token size and prefix-latent reference strategy for reference image (\textit{Ref.}). The first two rows are conducted without using sequence parallel. }
    \vspace{-0.2in}
    \label{tab:abl}
\end{table}

To investigate the roles of the proposed conditions, we examine several model variants, altering factors such as maximum token size, prefix latent strategy, text mask during training, and pose refinement during inference.


\noindent\textbf{Max Token Size.} Table~\ref{tab:abl} demonstrates the importance of generating long video sequences. Reducing the maximum token count from 480K (up to 249 frames) to 80K (about 13-41 frames) significantly decreases the model's temporal visibility, which affects its capacity to fully understand coherent videos. Moreover, when generating longer videos with fewer tokens, the model requires more iterations that rely on the last frame of the previous sequence, which increases the likelihood of accumulated errors from earlier frames. Larger token count is still essential for better video performance while comparing to token size 400K. 

\noindent\textbf{Prefix-Latent Reference Strategy.} In Table~\ref{tab:abl}, we also compare two reference strategies, prefix-latent reference and latent concatenation. The results illustrates that while both methods perform comparably on single image generation, while concatenate latent lacks fidelity when applied to video generation. Additionally, the concatenate latent introduces extra computational cost, making it less efficient.

\noindent\textbf{Pose Refinement.} Relying on the pose adapter may lead to size variations or scaling inconsistencies in the hands and face, as shown in Figure~\ref{fig:abl}. Integrating pose adapter and Keypoint-DiT for pose refinement effectively resolves these inconsistencies, ensuring more accurate alignment with the reference image.


\subsection{User Study}

\begin{table}[t]
    \centering
    \footnotesize
    \begin{tabular}{c|cccc}
    \Xhline{1pt}
        Method & \textit{A.A.} & \textit{M.M.} & Ours & GT \\ \hline
        Consistency$\uparrow$ & 42\% / 35\% & 66\% / 79\% & \cellcolor[HTML]{D9D9D9}{98\% / 98\%}  & 99\% / - \\
        Preservation$\uparrow$ & 65\% / 71\% & 69\% / 68\% & \cellcolor[HTML]{D9D9D9}{99\%} / 96\% & 99\% / - \\
        Quality$\uparrow$  & 72\% / 64\% & 79\% / 66\% & \cellcolor[HTML]{D9D9D9}{98\%} / 95\% & 99\% / - \\\Xhline{1pt}
    \end{tabular}
    \vspace{-0.1in}
    \caption{User study. The values represent the win rates of the current method compared to other methods. We ask users to rate the generated videos on our test set (left) and wild pose sequences with pose transfer (right), comparing with Animate Anyone (\textit{A.A.})~\cite{hu2024animate} and MimicMotion (\textit{M.M.})~\cite{zhang2024mimicmotion} in terms of temporal consistency, identify preservation, and visual quality.}
    \vspace{-0.2in}
    \label{tab:user}
\end{table}

To assess the quality of our method and state-of-the-art approaches from a human perspective, we conduct a blind user study with 10 participants. As summarized in Table~\ref{tab:user}, our method significantly outperforms others in terms of temporal consistency, identity preservation, and visual quality, highlighting its effectiveness. Details in Appendix.C.
\section{Conclusion}
\label{sec:conclusion}

In conclusion, we introduce HumanDiT, a pose-guided DiT-based framework designed for high-fidelity, long-form human motion video generation, trained on a large-scale human video dataset. HumanDiT supports variable resolutions and sequence lengths, and ensures visual consistency through a prefix-latent reference strategy. Leveraging poses generated by the proposed Keypoint-DiT and pose adapter, HumanDiT produces high-quality, pose-consistent videos with wild images and poses. Extensive experiments demonstrate its superior performance across diverse scenarios.

{
    \small
    \bibliographystyle{ieeenat_fullname}
    \bibliography{main}

\begin{thebibliography}{73}
\providecommand{\natexlab}[1]{#1}
\providecommand{\url}[1]{\texttt{#1}}
\expandafter\ifx\csname urlstyle\endcsname\relax
  \providecommand{\doi}[1]{doi: #1}\else
  \providecommand{\doi}{doi: \begingroup \urlstyle{rm}\Url}\fi

\bibitem[Balaji et~al.(2019)Balaji, Min, Bai, Chellappa, and Graf]{balaji2019conditional}
Yogesh Balaji, Martin~Renqiang Min, Bing Bai, Rama Chellappa, and Hans~Peter Graf.
\newblock Conditional gan with discriminative filter generation for text-to-video synthesis.
\newblock In \emph{IJCAI}, page~2, 2019.

\bibitem[Bar-Tal et~al.(2023)Bar-Tal, Yariv, Lipman, and Dekel]{bar2023multidiffusion}
Omer Bar-Tal, Lior Yariv, Yaron Lipman, and Tali Dekel.
\newblock Multidiffusion: Fusing diffusion paths for controlled image generation.
\newblock 2023.

\bibitem[Bar-Tal et~al.(2024)Bar-Tal, Chefer, Tov, Herrmann, Paiss, Zada, Ephrat, Hur, Li, Michaeli, et~al.]{bar2024lumiere}
Omer Bar-Tal, Hila Chefer, Omer Tov, Charles Herrmann, Roni Paiss, Shiran Zada, Ariel Ephrat, Junhwa Hur, Yuanzhen Li, Tomer Michaeli, et~al.
\newblock Lumiere: A space-time diffusion model for video generation.
\newblock \emph{arXiv preprint arXiv:2401.12945}, 2024.

\bibitem[Blackforestlabs(2024)]{flux}
Blackforestlabs.
\newblock Flux.
\newblock \url{https://github.com/black-forest-labs/flux}, 2024.

\bibitem[Blattmann et~al.(2023)Blattmann, Dockhorn, Kulal, Mendelevitch, Kilian, Lorenz, Levi, English, Voleti, Letts, et~al.]{blattmann2023stable}
Andreas Blattmann, Tim Dockhorn, Sumith Kulal, Daniel Mendelevitch, Maciej Kilian, Dominik Lorenz, Yam Levi, Zion English, Vikram Voleti, Adam Letts, et~al.
\newblock Stable video diffusion: Scaling latent video diffusion models to large datasets.
\newblock \emph{arXiv preprint arXiv:2311.15127}, 2023.

\bibitem[Cao et~al.(2017)Cao, Simon, Wei, and Sheikh]{cao2017realtime}
Zhe Cao, Tomas Simon, Shih-En Wei, and Yaser Sheikh.
\newblock Realtime multi-person 2d pose estimation using part affinity fields.
\newblock In \emph{CVPR}, pages 7291--7299, 2017.

\bibitem[Chan et~al.(2019)Chan, Ginosar, Zhou, and Efros]{chan2019everybody}
Caroline Chan, Shiry Ginosar, Tinghui Zhou, and Alexei~A Efros.
\newblock Everybody dance now.
\newblock In \emph{ICCV}, pages 5933--5942, 2019.

\bibitem[Chang et~al.(2023)Chang, Shi, Gao, Fu, Xu, Song, Yan, Yang, and Soleymani]{chang2023magicdance}
Di Chang, Yichun Shi, Quankai Gao, Jessica Fu, Hongyi Xu, Guoxian Song, Qing Yan, Xiao Yang, and Mohammad Soleymani.
\newblock Magicdance: Realistic human dance video generation with motions \& facial expressions transfer.
\newblock \emph{arXiv preprint arXiv:2311.12052}, 2023.

\bibitem[Ge et~al.(2022)Ge, Hayes, Yang, Yin, Pang, Jacobs, Huang, and Parikh]{ge2022long}
Songwei Ge, Thomas Hayes, Harry Yang, Xi Yin, Guan Pang, David Jacobs, Jia-Bin Huang, and Devi Parikh.
\newblock Long video generation with time-agnostic vqgan and time-sensitive transformer.
\newblock In \emph{ECCV}, pages 102--118, 2022.

\bibitem[Ginosar et~al.(2019)Ginosar, Bar, Kohavi, Chan, Owens, and Malik]{ginosar2019learning}
Shiry Ginosar, Amir Bar, Gefen Kohavi, Caroline Chan, Andrew Owens, and Jitendra Malik.
\newblock Learning individual styles of conversational gesture.
\newblock In \emph{CVPR}, pages 3497--3506, 2019.

\bibitem[Goodfellow et~al.(2014)Goodfellow, Pouget-Abadie, Mirza, Xu, Warde-Farley, Ozair, Courville, and Bengio]{goodfellow2014generative}
Ian Goodfellow, Jean Pouget-Abadie, Mehdi Mirza, Bing Xu, David Warde-Farley, Sherjil Ozair, Aaron Courville, and Yoshua Bengio.
\newblock Generative adversarial nets.
\newblock \emph{NeurIPS}, 27, 2014.

\bibitem[Guan et~al.(2024)Guan, Yang, Wang, Zhou, He, Xu, Feng, Ding, Wang, Xie, et~al.]{guan2024talk}
Jiazhi Guan, Quanwei Yang, Kaisiyuan Wang, Hang Zhou, Shengyi He, Zhiliang Xu, Haocheng Feng, Errui Ding, Jingdong Wang, Hongtao Xie, et~al.
\newblock Talk-act: Enhance textural-awareness for 2d speaking avatar reenactment with diffusion model.
\newblock In \emph{SIGGRAPH Asia}, pages 1--11, 2024.

\bibitem[Guo et~al.(2023)Guo, Yang, Rao, Liang, Wang, Qiao, Agrawala, Lin, and Dai]{guo2023animatediff}
Yuwei Guo, Ceyuan Yang, Anyi Rao, Zhengyang Liang, Yaohui Wang, Yu Qiao, Maneesh Agrawala, Dahua Lin, and Bo Dai.
\newblock Animatediff: Animate your personalized text-to-image diffusion models without specific tuning.
\newblock \emph{arXiv preprint arXiv:2307.04725}, 2023.

\bibitem[He et~al.(2016)He, Zhang, Ren, and Sun]{he2016deep}
Kaiming He, Xiangyu Zhang, Shaoqing Ren, and Jian Sun.
\newblock Deep residual learning for image recognition.
\newblock In \emph{CVPR}, pages 770--778, 2016.

\bibitem[Heusel et~al.(2017)Heusel, Ramsauer, Unterthiner, Nessler, and Hochreiter]{heusel2017gans}
Martin Heusel, Hubert Ramsauer, Thomas Unterthiner, Bernhard Nessler, and Sepp Hochreiter.
\newblock Gans trained by a two time-scale update rule converge to a local nash equilibrium.
\newblock \emph{NeurIPS}, 30, 2017.

\bibitem[Ho et~al.(2020)Ho, Jain, and Abbeel]{ho2020denoising}
Jonathan Ho, Ajay Jain, and Pieter Abbeel.
\newblock Denoising diffusion probabilistic models.
\newblock \emph{NeurIPS}, 33:\penalty0 6840--6851, 2020.

\bibitem[Hore and Ziou(2010)]{hore2010image}
Alain Hore and Djemel Ziou.
\newblock Image quality metrics: Psnr vs. ssim.
\newblock In \emph{ICPR}, pages 2366--2369. IEEE, 2010.

\bibitem[Hu(2024)]{hu2024animate}
Li Hu.
\newblock Animate anyone: Consistent and controllable image-to-video synthesis for character animation.
\newblock In \emph{CVPR}, pages 8153--8163, 2024.

\bibitem[Jafarian and Park(2021)]{jafarian2021learning}
Yasamin Jafarian and Hyun~Soo Park.
\newblock Learning high fidelity depths of dressed humans by watching social media dance videos.
\newblock In \emph{CVPR}, pages 12753--12762, 2021.

\bibitem[Jocher et~al.(2023)Jocher, Qiu, and Chaurasia]{Jocher_Ultralytics_YOLO_2023}
Glenn Jocher, Jing Qiu, and Ayush Chaurasia.
\newblock Ultralytics yolo.
\newblock \url{https://github.com/ultralytics/ultralytics}, 2023.

\bibitem[Karaev et~al.(2023)Karaev, Rocco, Graham, Neverova, Vedaldi, and Rupprecht]{karaev2023cotracker}
Nikita Karaev, Ignacio Rocco, Benjamin Graham, Natalia Neverova, Andrea Vedaldi, and Christian Rupprecht.
\newblock Cotracker: It is better to track together.
\newblock \emph{arXiv preprint arXiv:2307.07635}, 2023.

\bibitem[Khirodkar et~al.(2024)Khirodkar, Bagautdinov, Martinez, Zhaoen, James, Selednik, Anderson, and Saito]{khirodkar2024sapiens}
Rawal Khirodkar, Timur Bagautdinov, Julieta Martinez, Su Zhaoen, Austin James, Peter Selednik, Stuart Anderson, and Shunsuke Saito.
\newblock Sapiens: Foundation for human vision models.
\newblock \emph{arXiv preprint arXiv:2408.12569}, 2024.

\bibitem[Kingma(2013)]{kingma2013auto}
Diederik~P Kingma.
\newblock Auto-encoding variational bayes.
\newblock \emph{arXiv preprint arXiv:1312.6114}, 2013.

\bibitem[Li et~al.(2021)Li, Xue, Baranwal, Li, and You]{li2021sequence}
Shenggui Li, Fuzhao Xue, Chaitanya Baranwal, Yongbin Li, and Yang You.
\newblock Sequence parallelism: Long sequence training from system perspective.
\newblock \emph{arXiv preprint arXiv:2105.13120}, 2021.

\bibitem[Liao et~al.(2020)Liao, Zhang, Wang, Zhu, Zuo, and Yang]{liao2020speech2video}
Miao Liao, Sibo Zhang, Peng Wang, Hao Zhu, Xinxin Zuo, and Ruigang Yang.
\newblock Speech2video synthesis with 3d skeleton regularization and expressive body poses.
\newblock In \emph{ACCV}, 2020.

\bibitem[Lin et~al.(2024)Lin, Jiang, Liang, Zhong, Yang, and Zheng]{lin2024cyberhost}
Gaojie Lin, Jianwen Jiang, Chao Liang, Tianyun Zhong, Jiaqi Yang, and Yanbo Zheng.
\newblock Cyberhost: Taming audio-driven avatar diffusion model with region codebook attention.
\newblock \emph{arXiv preprint arXiv:2409.01876}, 2024.

\bibitem[Liu et~al.(2024)Liu, Yang, Akiyama, Huang, Li, Kuriyama, and Taketomi]{liu2024tango}
Haiyang Liu, Xingchao Yang, Tomoya Akiyama, Yuantian Huang, Qiaoge Li, Shigeru Kuriyama, and Takafumi Taketomi.
\newblock Tango: Co-speech gesture video reenactment with hierarchical audio motion embedding and diffusion interpolation.
\newblock \emph{arXiv preprint arXiv:2410.04221}, 2024.

\bibitem[Liu et~al.(2019)Liu, Dou, Chen, Qin, and Heng]{liu2019multi}
Lihao Liu, Qi Dou, Hao Chen, Jing Qin, and Pheng-Ann Heng.
\newblock Multi-task deep model with margin ranking loss for lung nodule analysis.
\newblock \emph{IEEE transactions on medical imaging}, 39\penalty0 (3):\penalty0 718--728, 2019.

\bibitem[Loper et~al.(2023)Loper, Mahmood, Romero, Pons-Moll, and Black]{loper2023smpl}
Matthew Loper, Naureen Mahmood, Javier Romero, Gerard Pons-Moll, and Michael~J Black.
\newblock Smpl: A skinned multi-person linear model.
\newblock In \emph{Seminal Graphics Papers: Pushing the Boundaries, Volume 2}, pages 851--866. 2023.

\bibitem[Lu et~al.(2024)Lu, Xu, Zhang, Wang, and Tao]{lu2024handrefiner}
Wenquan Lu, Yufei Xu, Jing Zhang, Chaoyue Wang, and Dacheng Tao.
\newblock Handrefiner: Refining malformed hands in generated images by diffusion-based conditional inpainting.
\newblock In \emph{ACM MM}, pages 7085--7093, 2024.

\bibitem[Mildenhall et~al.(2021)Mildenhall, Srinivasan, Tancik, Barron, Ramamoorthi, and Ng]{mildenhall2021nerf}
Ben Mildenhall, Pratul~P Srinivasan, Matthew Tancik, Jonathan~T Barron, Ravi Ramamoorthi, and Ren Ng.
\newblock Nerf: Representing scenes as neural radiance fields for view synthesis.
\newblock \emph{Communications of the ACM}, 65\penalty0 (1):\penalty0 99--106, 2021.

\bibitem[MooreThreads(2024)]{moore}
MooreThreads.
\newblock Moore-animateanyone.
\newblock \url{https://github.com/MooreThreads/Moore-AnimateAnyone}, 2024.

\bibitem[Ni et~al.(2023)Ni, Shi, Li, Huang, and Min]{ni2023conditional}
Haomiao Ni, Changhao Shi, Kai Li, Sharon~X Huang, and Martin~Renqiang Min.
\newblock Conditional image-to-video generation with latent flow diffusion models.
\newblock In \emph{CVPR}, pages 18444--18455, 2023.

\bibitem[OpenAI(2024)]{openai2024sora}
OpenAI.
\newblock Sora: Creating video from text.
\newblock \url{https://openai.com/sora}, 2024.
\newblock Accessed: 2024-10-13.

\bibitem[PaddlePaddle(2021)]{PaddleOCR}
PaddlePaddle.
\newblock Paddleocr.
\newblock \url{https://github.com/PaddlePaddle/PaddleOCR}, 2021.

\bibitem[Peebles and Xie(2023)]{peebles2023scalable}
William Peebles and Saining Xie.
\newblock Scalable diffusion models with transformers.
\newblock In \emph{ICCV}, pages 4195--4205, 2023.

\bibitem[Ren et~al.(2020)Ren, Li, Liu, and Li]{ren2020deep}
Yurui Ren, Ge Li, Shan Liu, and Thomas~H Li.
\newblock Deep spatial transformation for pose-guided person image generation and animation.
\newblock \emph{TIP}, 29:\penalty0 8622--8635, 2020.

\bibitem[Rombach et~al.(2022)Rombach, Blattmann, Lorenz, Esser, and Ommer]{rombach2022high}
Robin Rombach, Andreas Blattmann, Dominik Lorenz, Patrick Esser, and Bj{\"o}rn Ommer.
\newblock High-resolution image synthesis with latent diffusion models.
\newblock In \emph{CVPR}, pages 10684--10695, 2022.

\bibitem[Ronneberger et~al.(2015)Ronneberger, Fischer, and Brox]{ronneberger2015u}
Olaf Ronneberger, Philipp Fischer, and Thomas Brox.
\newblock U-net: Convolutional networks for biomedical image segmentation.
\newblock In \emph{Medical image computing and computer-assisted intervention--MICCAI 2015: 18th international conference, Munich, Germany, October 5-9, 2015, proceedings, part III 18}, pages 234--241. Springer, 2015.

\bibitem[Shao et~al.(2024)Shao, Pang, Zheng, Sun, and Liu]{shao2024human4dit}
Ruizhi Shao, Youxin Pang, Zerong Zheng, Jingxiang Sun, and Yebin Liu.
\newblock Human4dit: 360-degree human video generation with 4d diffusion transformer.
\newblock \emph{TOG}, 43\penalty0 (6), 2024.

\bibitem[Siarohin et~al.(2019)Siarohin, Lathuili{\`e}re, Tulyakov, Ricci, and Sebe]{siarohin2019animating}
Aliaksandr Siarohin, St{\'e}phane Lathuili{\`e}re, Sergey Tulyakov, Elisa Ricci, and Nicu Sebe.
\newblock Animating arbitrary objects via deep motion transfer.
\newblock In \emph{CVPR}, pages 2377--2386, 2019.

\bibitem[Siarohin et~al.(2021)Siarohin, Woodford, Ren, Chai, and Tulyakov]{siarohin2021motion}
Aliaksandr Siarohin, Oliver~J Woodford, Jian Ren, Menglei Chai, and Sergey Tulyakov.
\newblock Motion representations for articulated animation.
\newblock In \emph{CVPR}, pages 13653--13662, 2021.

\bibitem[Singer et~al.(2022)Singer, Polyak, Hayes, Yin, An, Zhang, Hu, Yang, Ashual, Gafni, et~al.]{singer2022make}
Uriel Singer, Adam Polyak, Thomas Hayes, Xi Yin, Jie An, Songyang Zhang, Qiyuan Hu, Harry Yang, Oron Ashual, Oran Gafni, et~al.
\newblock Make-a-video: Text-to-video generation without text-video data.
\newblock \emph{arXiv preprint arXiv:2209.14792}, 2022.

\bibitem[Song et~al.(2020)Song, Meng, and Ermon]{song2020denoising}
Jiaming Song, Chenlin Meng, and Stefano Ermon.
\newblock Denoising diffusion implicit models.
\newblock \emph{arXiv preprint arXiv:2010.02502}, 2020.

\bibitem[Su et~al.(2024)Su, Ahmed, Lu, Pan, Bo, and Liu]{su2024roformer}
Jianlin Su, Murtadha Ahmed, Yu Lu, Shengfeng Pan, Wen Bo, and Yunfeng Liu.
\newblock Roformer: Enhanced transformer with rotary position embedding.
\newblock \emph{Neurocomputing}, 568:\penalty0 127063, 2024.

\bibitem[Sun et~al.(2024)Sun, Jiang, Chen, Zhang, Peng, Luo, and Yuan]{sun2024autoregressive}
Peize Sun, Yi Jiang, Shoufa Chen, Shilong Zhang, Bingyue Peng, Ping Luo, and Zehuan Yuan.
\newblock Autoregressive model beats diffusion: Llama for scalable image generation.
\newblock \emph{arXiv preprint arXiv:2406.06525}, 2024.

\bibitem[Tan et~al.(2024)Tan, Gong, Wang, Zhang, Zheng, Zheng, Zheng, Chen, and Yang]{tan2024animate}
Shuai Tan, Biao Gong, Xiang Wang, Shiwei Zhang, Dandan Zheng, Ruobing Zheng, Kecheng Zheng, Jingdong Chen, and Ming Yang.
\newblock Animate-x: Universal character image animation with enhanced motion representation.
\newblock \emph{arXiv preprint arXiv:2410.10306}, 2024.

\bibitem[Unterthiner et~al.(2018)Unterthiner, Van~Steenkiste, Kurach, Marinier, Michalski, and Gelly]{unterthiner2018towards}
Thomas Unterthiner, Sjoerd Van~Steenkiste, Karol Kurach, Raphael Marinier, Marcin Michalski, and Sylvain Gelly.
\newblock Towards accurate generative models of video: A new metric \& challenges.
\newblock \emph{arXiv preprint arXiv:1812.01717}, 2018.

\bibitem[Vaswani(2017)]{vaswani2017attention}
A Vaswani.
\newblock Attention is all you need.
\newblock \emph{NeurIPS}, 2017.

\bibitem[Vondrick et~al.(2016)Vondrick, Pirsiavash, and Torralba]{vondrick2016generating}
Carl Vondrick, Hamed Pirsiavash, and Antonio Torralba.
\newblock Generating videos with scene dynamics.
\newblock \emph{NeurIPS}, 29, 2016.

\bibitem[Wang et~al.(2023)Wang, Gu, Hu, Xu, Xu, and Liang]{wang2023dreamvideo}
Cong Wang, Jiaxi Gu, Panwen Hu, Songcen Xu, Hang Xu, and Xiaodan Liang.
\newblock Dreamvideo: High-fidelity image-to-video generation with image retention and text guidance.
\newblock \emph{arXiv preprint arXiv:2312.03018}, 2023.

\bibitem[Wang et~al.(2024{\natexlab{a}})Wang, Li, Lin, Zhai, Lin, Yang, Zhang, Liu, and Wang]{wang2024disco}
Tan Wang, Linjie Li, Kevin Lin, Yuanhao Zhai, Chung-Ching Lin, Zhengyuan Yang, Hanwang Zhang, Zicheng Liu, and Lijuan Wang.
\newblock Disco: Disentangled control for realistic human dance generation.
\newblock In \emph{CVPR}, pages 9326--9336, 2024{\natexlab{a}}.

\bibitem[Wang et~al.(2021)Wang, Mallya, and Liu]{wang2021one}
Ting-Chun Wang, Arun Mallya, and Ming-Yu Liu.
\newblock One-shot free-view neural talking-head synthesis for video conferencing.
\newblock In \emph{CVPR}, pages 10039--10049, 2021.

\bibitem[Wang et~al.(2024{\natexlab{b}})Wang, Zhang, Gao, Wang, Zhou, Zhang, Yan, and Sang]{wang2024unianimate}
Xiang Wang, Shiwei Zhang, Changxin Gao, Jiayu Wang, Xiaoqiang Zhou, Yingya Zhang, Luxin Yan, and Nong Sang.
\newblock Unianimate: Taming unified video diffusion models for consistent human image animation.
\newblock \emph{arXiv preprint arXiv:2406.01188}, 2024{\natexlab{b}}.

\bibitem[Wang et~al.(2004)Wang, Bovik, Sheikh, and Simoncelli]{wang2004image}
Zhou Wang, Alan~C Bovik, Hamid~R Sheikh, and Eero~P Simoncelli.
\newblock Image quality assessment: from error visibility to structural similarity.
\newblock \emph{TIP}, 13\penalty0 (4):\penalty0 600--612, 2004.

\bibitem[Weng et~al.(2022)Weng, Curless, Srinivasan, Barron, and Kemelmacher-Shlizerman]{weng2022humannerf}
Chung-Yi Weng, Brian Curless, Pratul~P Srinivasan, Jonathan~T Barron, and Ira Kemelmacher-Shlizerman.
\newblock Humannerf: Free-viewpoint rendering of moving people from monocular video.
\newblock In \emph{CVPR}, pages 16210--16220, 2022.

\bibitem[Wu et~al.(2023)Wu, Zhang, Zhang, Chen, Li, Liao, Wang, Zhang, Sun, Yan, Min, Zhai, and Lin]{wu2023qalign}
Haoning Wu, Zicheng Zhang, Weixia Zhang, Chaofeng Chen, Chunyi Li, Liang Liao, Annan Wang, Erli Zhang, Wenxiu Sun, Qiong Yan, Xiongkuo Min, Guangtai Zhai, and Weisi Lin.
\newblock Q-align: Teaching lmms for visual scoring via discrete text-defined levels.
\newblock \emph{arXiv preprint arXiv:2312.17090}, 2023.

\bibitem[Xu et~al.(2016)Xu, Mei, Yao, and Rui]{xu2016msr}
Jun Xu, Tao Mei, Ting Yao, and Yong Rui.
\newblock Msr-vtt: A large video description dataset for bridging video and language.
\newblock In \emph{CVPR}, pages 5288--5296, 2016.

\bibitem[Xu et~al.(2024{\natexlab{a}})Xu, Zou, Huang, Chen, Liu, Cheng, Shi, and Huang]{xu2024easyanimate}
Jiaqi Xu, Xinyi Zou, Kunzhe Huang, Yunkuo Chen, Bo Liu, MengLi Cheng, Xing Shi, and Jun Huang.
\newblock Easyanimate: A high-performance long video generation method based on transformer architecture.
\newblock \emph{arXiv preprint arXiv:2405.18991}, 2024{\natexlab{a}}.

\bibitem[Xu et~al.(2024{\natexlab{b}})Xu, Zhang, Liew, Yan, Liu, Zhang, Feng, and Shou]{xu2024magicanimate}
Zhongcong Xu, Jianfeng Zhang, Jun~Hao Liew, Hanshu Yan, Jia-Wei Liu, Chenxu Zhang, Jiashi Feng, and Mike~Zheng Shou.
\newblock Magicanimate: Temporally consistent human image animation using diffusion model.
\newblock In \emph{CVPR}, pages 1481--1490, 2024{\natexlab{b}}.

\bibitem[Xue et~al.(2024)Xue, Wang, Tian, Ma, Wang, Zhao, Min, Zhao, Zhang, Shum, et~al.]{xue2024follow}
Jingyun Xue, Hongfa Wang, Qi Tian, Yue Ma, Andong Wang, Zhiyuan Zhao, Shaobo Min, Wenzhe Zhao, Kaihao Zhang, Heung-Yeung Shum, et~al.
\newblock Follow-your-pose v2: Multiple-condition guided character image animation for stable pose control.
\newblock \emph{arXiv preprint arXiv:2406.03035}, 2024.

\bibitem[Yang et~al.()Yang, Guan, Wang, Yu, Chu, Zhou, Feng, Feng, Ding, Wang, et~al.]{yangshowmaker}
Quanwei Yang, Jiazhi Guan, Kaisiyuan Wang, Lingyun Yu, Wenqing Chu, Hang Zhou, ZhiQiang Feng, Haocheng Feng, Errui Ding, Jingdong Wang, et~al.
\newblock Showmaker: Creating high-fidelity 2d human video via fine-grained diffusion modeling.
\newblock In \emph{NeurIPS}.

\bibitem[Yang et~al.(2023)Yang, Zeng, Yuan, and Li]{yang2023effective}
Zhendong Yang, Ailing Zeng, Chun Yuan, and Yu Li.
\newblock Effective whole-body pose estimation with two-stages distillation.
\newblock In \emph{ICCV}, pages 4210--4220, 2023.

\bibitem[Yang et~al.(2024)Yang, Teng, Zheng, Ding, Huang, Xu, Yang, Hong, Zhang, Feng, et~al.]{yang2024cogvideox}
Zhuoyi Yang, Jiayan Teng, Wendi Zheng, Ming Ding, Shiyu Huang, Jiazheng Xu, Yuanming Yang, Wenyi Hong, Xiaohan Zhang, Guanyu Feng, et~al.
\newblock Cogvideox: Text-to-video diffusion models with an expert transformer.
\newblock \emph{arXiv preprint arXiv:2408.06072}, 2024.

\bibitem[Ye et~al.(2023)Ye, Jiang, Ren, Liu, He, and Zhao]{ye2023geneface}
Zhenhui Ye, Ziyue Jiang, Yi Ren, Jinglin Liu, Jinzheng He, and Zhou Zhao.
\newblock Geneface: Generalized and high-fidelity audio-driven 3d talking face synthesis.
\newblock \emph{arXiv preprint arXiv:2301.13430}, 2023.

\bibitem[Zeng et~al.(2024)Zeng, Wei, Zheng, Zou, Wei, Zhang, and Li]{zeng2024make}
Yan Zeng, Guoqiang Wei, Jiani Zheng, Jiaxin Zou, Yang Wei, Yuchen Zhang, and Hang Li.
\newblock Make pixels dance: High-dynamic video generation.
\newblock In \emph{CVPR}, pages 8850--8860, 2024.

\bibitem[Zhang et~al.(2023)Zhang, Rao, and Agrawala]{zhang2023adding}
Lvmin Zhang, Anyi Rao, and Maneesh Agrawala.
\newblock Adding conditional control to text-to-image diffusion models.
\newblock In \emph{ICCV}, pages 3836--3847, 2023.

\bibitem[Zhang et~al.(2018)Zhang, Isola, Efros, Shechtman, and Wang]{zhang2018unreasonable}
Richard Zhang, Phillip Isola, Alexei~A Efros, Eli Shechtman, and Oliver Wang.
\newblock The unreasonable effectiveness of deep features as a perceptual metric.
\newblock In \emph{CVPR}, pages 586--595, 2018.

\bibitem[Zhang et~al.(2024{\natexlab{a}})Zhang, Gu, Wang, Wang, Cheng, Zhu, and Zou]{zhang2024mimicmotion}
Yuang Zhang, Jiaxi Gu, Li-Wen Wang, Han Wang, Junqi Cheng, Yuefeng Zhu, and Fangyuan Zou.
\newblock Mimicmotion: High-quality human motion video generation with confidence-aware pose guidance.
\newblock \emph{arXiv preprint arXiv:2406.19680}, 2024{\natexlab{a}}.

\bibitem[Zhang et~al.(2024{\natexlab{b}})Zhang, Liao, Li, Qin, and Wang]{zhang2024tora}
Zhenghao Zhang, Junchao Liao, Menghao Li, Long Qin, and Weizhi Wang.
\newblock Tora: Trajectory-oriented diffusion transformer for video generation.
\newblock \emph{arXiv preprint arXiv:2407.21705}, 2024{\natexlab{b}}.

\bibitem[Zhou et~al.(2022)Zhou, Wang, Yan, Lv, Zhu, and Feng]{zhou2022magicvideo}
Daquan Zhou, Weimin Wang, Hanshu Yan, Weiwei Lv, Yizhe Zhu, and Jiashi Feng.
\newblock Magicvideo: Efficient video generation with latent diffusion models.
\newblock \emph{arXiv preprint arXiv:2211.11018}, 2022.

\bibitem[Zhou et~al.(2024)Zhou, Wang, Chen, Bai, Li, Zhang, Xu, Yang, and Wang]{zhou2024realisdance}
Jingkai Zhou, Benzhi Wang, Weihua Chen, Jingqi Bai, Dongyang Li, Aixi Zhang, Hao Xu, Mingyang Yang, and Fan Wang.
\newblock Realisdance: Equip controllable character animation with realistic hands.
\newblock \emph{arXiv preprint arXiv:2409.06202}, 2024.

\bibitem[Zhu et~al.(2024)Zhu, Chen, Dai, Xu, Cao, Yao, Zhu, and Zhu]{zhu2024champ}
Shenhao Zhu, Junming~Leo Chen, Zuozhuo Dai, Yinghui Xu, Xun Cao, Yao Yao, Hao Zhu, and Siyu Zhu.
\newblock Champ: Controllable and consistent human image animation with 3d parametric guidance.
\newblock \emph{arXiv preprint arXiv:2403.14781}, 2024.

\end{thebibliography}
}

\clearpage
\setcounter{page}{1}
\maketitlesupplementary

\setcounter{figure}{0}
\setcounter{table}{0}
\setcounter{section}{0}
\renewcommand{\thefigure}{A\arabic{figure}}
\renewcommand{\thetable}{A\arabic{table}}
\renewcommand{\thesection}{\Alph{section}}

\section{More Details for Data Collection}
\label{sec:sup_data}

\subsection{Clarity Scoring Model}

\begin{figure}[htp]
    \centering
    \includegraphics[width=1.0\linewidth]{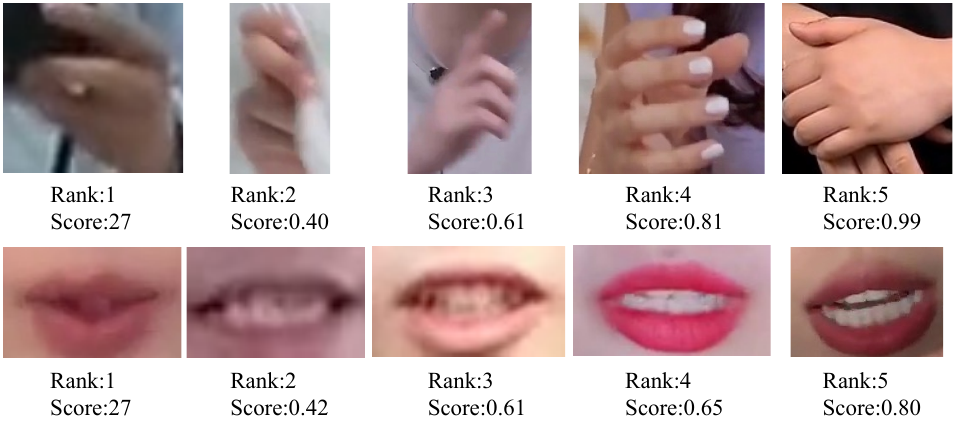}
    \vspace{-0.25in}
    \caption{The data samples and model results (score) of hand and teeth dataset. Ranks are the annotated clarity ascending orders. Scores are obtained from our clarity scoring model, respectively. }
    \vspace{-0.1in}
    \label{fig:scoremodel}
\end{figure}

In our video clarity assessment, we manually annotate 150,000 image sets, each comprising five images of teeth and hands extracted from various videos. These images are ranked based on clarity, occlusion level, and detail quality of the target region. The dataset is divided into training and testing subsets with a 9:1 ratio. We develop a lightweight clarity scoring model using ResNet-18~\cite{he2016deep} as the backbone, with output scores derived via a sigmoid function. During training, we randomly select image pairs from each set that differed by at least 2 points in clarity rank and apply MarginRankingLoss~\cite{liu2019multi} to compute the loss between these paired scores. The models for teeth and hand clarity are trained over 100 epochs, achieving test accuracies of 79.8\% and 83.5\%, respectively. These clarity scores are then used to filter high-quality videos, thereby enhancing the dataset's visual quality. Data samples and results are presented in Figure~\ref{fig:scoremodel}.

\subsection{Data Extraction and Preprocessing}
We developed a robust data extraction and filtering pipeline to efficiently isolate and retain high-quality human action videos from extensive in-the-wild datasets. This pipeline ensures the selection of high-clarity, visually consistent video segments appropriate for training and evaluation.

The processing of videos begins with human tracking followed by video cropping. Using YOLO~\cite{Jocher_Ultralytics_YOLO_2023}, we detect and track each human sequence, ensuring each segment contains a single subject. The cropping dimensions for each segment are dynamically adjusted based on the subject’s maximum motion range, preserving full-body visibility and enabling accurate pose analysis. The body masks can also be obtained by YOLO.

\subsection{Feature Extraction}
After identifying and segmenting each individual in the video, various key features were extracted, including human poses, background keypoints, and text bounding boxes.

\noindent1) \textbf{Pose Estimation.} Sapiens~\cite{khirodkar2024sapiens} is employed to obtain pose keypoints, providing robust human pose detection for each frame. Notably, to reduce redundant keypoints and avoid affecting the performance, some keypoints, such as the ears, are excluded from our model.

\noindent2) \textbf{Background Tracking} CoTracker~\cite{karaev2023cotracker} is applied to distinguish foreground movements from background changes, capturing the necessary spatial consistency for high-quality generation.

\noindent3) \textbf{Text Detection.} PaddleOCR~\cite{PaddleOCR} is used to identify and mark text regions in each frame, mitigating the potential interference of text artifacts with the generated data.

To ensure the inclusion of only high-quality frames, we evaluate each video segment at 5-frame intervals using both Q-Align~\cite{wu2023qalign} and custom scoring models. These regions of hand and teeth are obtained from the estimated keypoints. Frames with poor visibility or occlusion of these regions are excluded. This clarity scoring process ensures the final dataset contains samples with clearly visible and well-defined features. Furthermore, image histograms are computed to detect frames with excessive blur, visual transitions, or other quality degradations, further refining the visual consistency across samples.

\subsection{Filtering Criteria}
To ensure the high quality of the extracted videos, we implemented the following measures.

\noindent\textbf{Pose Quality.} A threshold of $0.3$ is set for the relative speed of keypoints to ensure reliable localization of key body parts. The relative speed is determined by measuring the displacement velocity of one end of the human skeleton with respect to the other.

\noindent\textbf{Frame Count and Visibility Constraints.} We filter segments with fewer than $30$ frames and require at least $80\%$ visibility of the upper body and hands across frames.

\noindent\textbf{Visual Quality.} A minimum average IQA score of $60$ (via Q-Align) is required to retain acceptable visual quality, with all frames exceeding $45$ to avoid frames with intermittent blur. We set minimum scores of $0.50$ for average hand clarity and $0.40$ for average teeth clarity to select videos with high detail quality. Cap the histogram variance at $0.3$ to exclude frames that are low-quality, excessively blurred, or contain numerous transitions.

\noindent\textbf{Motion Analysis.} To prevent abrupt scene transitions, the maximum speed threshold across joints is set at $0.3$, and the maximum mask Intersection over Union (IoU) threshold is set at 0.7 between frames to ensure smooth, continuous motion frames.

This meticulously crafted pipeline enables the selection of high-quality video segments distinguished by consistent motion, visual clarity, and pose alignment, thus providing a reliable dataset for robust human motion generation tasks.

\section{More Details for Keypoint-DiT}
\label{sec:sup_posegeneration}

Keypoint-DiT, similar to Video DiT, uses the pose from the first frame as a prefix pose and denoises the subsequent $t-1$ noise frames to generate the following motions. To equip the Keypoint-DiT module with the ability to refine and transition between poses in pose transfer tasks, we incorporate reference keypoints as conditional inputs embedded within the token sequences. This setup allows the model to learn smooth transitions between poses by gradually moving from an initial reference pose to a target reference pose.

To simulate the transition from a reference image pose to a desired target pose, we use keypoints extracted from real human motion data as target references. To encourage effective transition capabilities, the initial $\tau$ frames preceding the target pose are set to zero, allowing the model to progressively refine its generated keypoints and interpolate smoothly between states. During training, there is a 50\% probability that the GT keypoints are provided as conditional inputs in each frame; in the remaining cases, zero-padding is used. This helps the model to generalize by learning to generate realistic transitions both with and without explicit keypoint guidance. Our collected dataset, sampled at 25 fps, is used to train the model with consistent temporal and spatial resolution across frames. All input keypoints are normalized to a square image space (ranging from 0 to 1) to unify the model’s understanding of body proportions and to ensure a consistent spatial reference for each pose. The training process was conducted on 8 H100 GPUs over 1 million steps with a learning rate of 0.0001, allowing the model to achieve high fidelity in both pose transfer and motion generation across variable input conditions.

\section{User Study}

To evaluate the quality of our approach in comparison to state-of-the-art methods, we conducted a blind user study focused on three primary metrics: temporal consistency, identity preservation, and visual quality. Each of these aspects is critical in human motion video generation:

\noindent\textbf{Temporal Consistency.} This metric evaluates the smoothness of transitions between frames, ensuring there are no abrupt changes or flickering artifacts that disrupt motion flow in the video. A high temporal consistency score indicates that the video maintains natural, fluid movement without noticeable jumps.
  
\noindent\textbf{Identity Preservation.} This measures how well the generated video retains the unique characteristics of the subject (such as facial and bodily features) and the background scene across frames. Effective identity preservation ensures that the person and setting remain visually coherent and true to the original throughout the video sequence.

\noindent\textbf{Visual Quality.} This aspect assesses the overall image clarity and absence of blurriness or visual artifacts. High visual quality indicates that the video maintains sharpness, fine details, and avoids pixelation or degradation that could affect the realism of the generated video.

In this study, we curated a set of 30 videos for tasks involving video reconstruction from the first frame and pose transfer. Ten domain experts from diverse industries participated in the evaluation. Each expert was randomly presented with pairs of anonymized videos, generated by our method and existing methods, for direct comparison across three metrics.

\section{More Results}
\label{sec:moreres}

\subsection{Quantitative Evaluation}

\begin{table*}[htb]
\centering\small
\begin{tabular}{c|c|ccccc|cc}
\Xhline{1pt}
\noindent\textbf{Model}        & \noindent\textbf{Dataset} & \noindent\textbf{FID}  $\downarrow$   & \noindent\textbf{SSIM}   $\uparrow$ & \noindent\textbf{PSNR}   $\uparrow$  & \noindent\textbf{LPIPS}  $\downarrow$  & \noindent\textbf{L1}    $\downarrow$  & \noindent\textbf{FID-VID}  $\downarrow$  & \noindent\textbf{FVD}  $\downarrow$   \\ \hline
\multirow{4}{*}{DisCo~\cite{wang2024disco}}  & TikTok              & 131 & 0.664              & 13.9& 0.396              & 5.18E-05   & 125        & 858\\ 
       & Talking & 229 & 0.619              & 12.7& 0.454              & 7.05E-05   & 229        & 1848              \\ 
       & Dancing              & 219 & 0.660              & 12.8& 0.431              & 5.14E-05     & 195      & 2414              \\ 
       & \cellcolor[HTML]{F0F0F0} Total             &\cellcolor[HTML]{F0F0F0} 210 &\cellcolor[HTML]{F0F0F0} 0.636              & \cellcolor[HTML]{F0F0F0} 12.9&\cellcolor[HTML]{F0F0F0} 0.439             &\cellcolor[HTML]{F0F0F0} 6.29E-05     &\cellcolor[HTML]{F0F0F0} 203      &\cellcolor[HTML]{F0F0F0} 1808             \\ \hline
\multirow{4}{*}{MagicDance~\cite{chang2023magicdance}}            & TikTok              & 71.8& 0.748              & 17.4& 0.318              & 3.19E-05     & 65.8     & 437\\ 
       & Talking & 106 & 0.719              & 17.6& 0.325              & 3.78E-05  &  97.7       & 874\\ 
       & Dancing              & 117  & 0.740    & 16.8& 0.340              & 2.68E-05  &  103       & 1394              \\ 
       &\cellcolor[HTML]{F0F0F0} Total &\cellcolor[HTML]{F0F0F0} 103 &\cellcolor[HTML]{F0F0F0} 0.728              &\cellcolor[HTML]{F0F0F0} 17.4&\cellcolor[HTML]{F0F0F0} 0.327              & \cellcolor[HTML]{F0F0F0} 3.43E-05  &\cellcolor[HTML]{F0F0F0} 93.4         &\cellcolor[HTML]{F0F0F0} 918\\ \hline
\multirow{4}{*}{AnimateAnyone~\cite{hu2024animate}}          & TikTok              & 60.7        & 0.734    & 17.4  & 0.287              & 3.07E-05    & 60.0       & 453\\ 
       & Talking & 75.8& 0.755    & 19.0  & 0.254              & 2.79E-05        & 64.9   & 677\\ 
       & Dancing              & 102 & 0.697              & 16.6& 0.328              & 2.63E-05 & 84.3 & 1060              \\ 
       &\cellcolor[HTML]{F0F0F0} Total &\cellcolor[HTML]{F0F0F0} 79.6& \cellcolor[HTML]{F0F0F0} 0.737    &\cellcolor[HTML]{F0F0F0} \cellcolor[HTML]{F0F0F0} 18.1&\cellcolor[HTML]{F0F0F0} 0.279              &\cellcolor[HTML]{F0F0F0} 2.80E-05        &\cellcolor[HTML]{F0F0F0} 68.8   &\cellcolor[HTML]{F0F0F0} 730\\ \hline
\multirow{4}{*}{MimicMotion~\cite{zhang2024mimicmotion}}           & TikTok              & 58.8  & 0.761    & 18.4  & 0.253  & 3.22E-05 & 35.2 & 323   \\ 
       & Talking & 56.3  & 0.794    & 21.5  & 0.215  & 2.27E-05 & 36.1 & 440   \\ 
       & Dancing              & 94.3& 0.737              & 17.8& 0.286              & 3.73E-05   & 51.6        & 607\\ 
       &\cellcolor[HTML]{F0F0F0} Total &\cellcolor[HTML]{F0F0F0} 65.4& \cellcolor[HTML]{F0F0F0} 0.776              &\cellcolor[HTML]{F0F0F0} 20.1&\cellcolor[HTML]{F0F0F0} 0.238              & \cellcolor[HTML]{F0F0F0} 2.76E-05  &\cellcolor[HTML]{F0F0F0} 39.5         &\cellcolor[HTML]{F0F0F0} 458\\ \hline
\multirow{4}{*}{Ours}            & TikTok              & 41.6 & 0.815              & 20.5 & 0.220             & 2.17E-05     & 24.3     & 237\\ 
       & Talking & 41.3 & 0.862              & 24.6 & 0.155              & 1.40E-05  &  18.2       & 257\\ 
       & Dancing              & 76.3  & 0.791    & 19.7 & 0.240              & 2.26E-05  &  45.4       & 548              \\ 
       & \cellcolor[HTML]{D9D9D9} Total& \cellcolor[HTML]{D9D9D9} 49.4 & \cellcolor[HTML]{D9D9D9} 0.838              & \cellcolor[HTML]{D9D9D9} 22.8  & \cellcolor[HTML]{D9D9D9} 0.186            & \cellcolor[HTML]{D9D9D9} 1.73E-05  & \cellcolor[HTML]{D9D9D9} 25.5         & \cellcolor[HTML]{D9D9D9} 320\\ \Xhline{1pt}

\end{tabular}
\vspace{-0.1in}
\caption{Quantitative comparison on the TikTok dataset~\cite{jafarian2021learning} and our self-collected talking and dancing test datasets with existing pose-guided body reenactment methods. }
\vspace{-0.1in}
\label{tab:metrics-total}
\end{table*}

Figure~\ref{tab:metrics-total} illustrates the performance of various methods across different scenarios, including the TikTok~\cite{jafarian2021learning} test set and our collected datasets of speech and dance videos. The dance dataset presents significant challenges due to its extensive motion amplitudes, in contrast to the speech dataset, which generally involves smaller movements and fewer variations, primarily concentrating on facial movements. The TikTok test set is comparatively simpler and fails to adequately assess each model's generalization capability in managing complex dance generation.

\subsection{More Visualization Results}

\begin{figure*}[htp]
    \centering
    \includegraphics[width=1.0\linewidth]{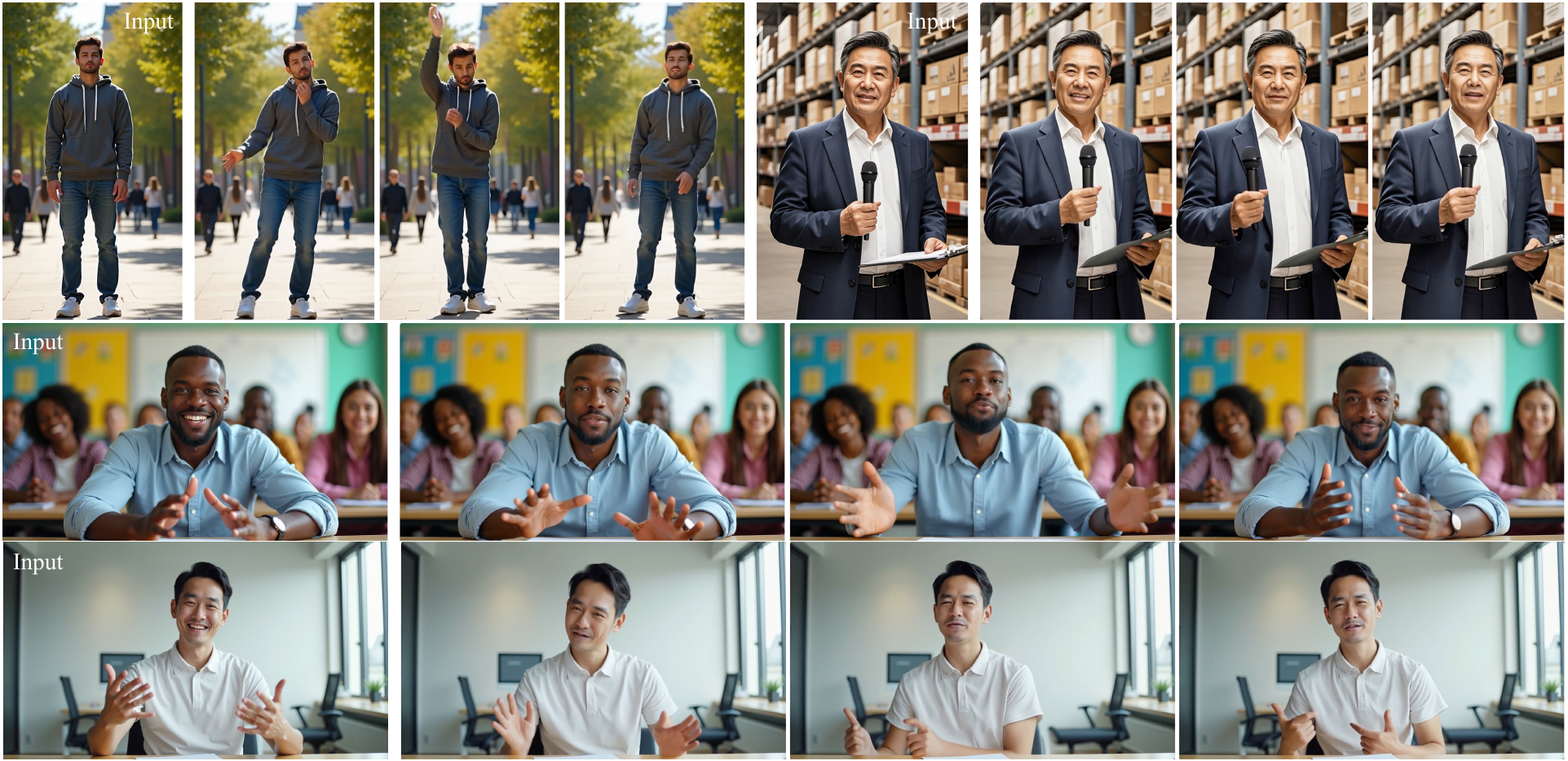}
    \vspace{-0.25in}
    \caption{More visualization results of video continuation.}
    \vspace{-0.25in}
    \label{fig:sup_gene}
\end{figure*}

\noindent\textbf{Video Continuation.} In Figure~\ref{fig:sup_gene}, we illustrate the video continuation capability of HumanDiT, showcasing its ability to generate realistic and coherent videos by extending motions based on the poses of given images across diverse scenarios. Specifically, our method excels in both full-body motion scenarios, such as dynamic dance sequences, and upper-body-focused tasks, such as speech gestures. These distinct tasks highlight the flexibility of HumanDiT to adapt to varying motion ranges and visual contexts while maintaining temporal consistency, identity preservation, and high rendering quality. 

\section{Limitations and Future Work.}

Despite promising results, HumanDiT has several limitations that can be addressed in future work. While the model excels in pose-guided human motion video generation, challenges remain in handling extreme body proportion variations and intricate hand and facial movements, which may affect visual consistency when the reference image and input pose are misaligned. Additionally, although HumanDiT generates long-form videos using a prefix-latent strategy, multi-batch inference may still result in slight error propagation, making it challenging for subsequent batches to fully preserve the identity characteristics of the first frame. Furthermore, generating videos at higher resolutions or longer durations still incurs high computational costs. Future work could explore more efficient architectures or hardware-accelerated optimizations to mitigate this issue. Lastly, the current pose generation model is limited by its inability to generate poses from images, speech, or textual descriptions. We plan to expand the model’s capabilities to incorporate these input modalities, enabling more dynamic pose generation based on a wider range of input features.

\end{document}